\documentclass[journal]{IEEEtran}

\IEEEoverridecommandlockouts                              

\usepackage{graphics} 
\usepackage{graphicx}
\usepackage{lettrine}
\usepackage{amsmath} 
\usepackage{amssymb}  
\usepackage{cite}
\usepackage{censor}
\usepackage{url}

\StopCensoring

\DeclareMathOperator{\sgn}{sgn}

\title{
System Identification of Admittance Models for Large Real-World Objects
}

\author{Nathan I. Baum$^{1}$, Nathaniel G. Luttmer, Mark A. Minor$^{1}$, \emph{IEEE Senior Member}
\thanks{This work was supported by the National Science Foundation under Grant No. 1911194.}%
\thanks{$^{1}$N. Baum and M. Minor are with the Department of Mechanical Engineering, University of Utah, Salt Lake City, UT 84112, USA. {\tt\small nate.baum@utah.edu, mark.minor@utah.edu}}%
\thanks{Code, data, and detailed derivations available \cite{baum2026admittance}}%
}

\begin{document}

\markboth{IEEE Robotics and Automation Letters, Vol. X, No. X, Month Year}
{Baum \MakeLowercase{\textit{et al.}}: System Identification of Admittance Models for Large Real-World Objects}

\maketitle

\begin{abstract}

Simulation of admittance-type models requires physically consistent dynamic models that are rarely available for off-the-shelf, everyday objects, limiting the fidelity of haptic interfaces that rely on such simulations. This paper presents the first complete workflow for producing physically consistent models of large real-world objects with various constraints and mechanisms, guaranteeing physical consistency of inertia and friction parameters. The workflow separates each object and identifies the \textit{handle} and \textit{body} in two stages, requiring no torque sensors at hinges, axles, or other constrained joints. Models are produced for a heavy, closer-actuated door and a wheelbarrow, representing objects of differing constraint types and model complexity. The door is modeled using four-bar linkage kinematics and a fluid dynamics-based lumped parameter model including opening, backcheck, swing, and latch zones. The wheelbarrow is modeled as a rigid body with a spherical wheel and no slip during rolling. Handle estimation RMS errors were below 0.64 N and 0.042 Nm across both objects. Door body estimation had RMS error of 2.19 Nm and wheelbarrow body estimation had RMS error of 6.77 Nm.
\end{abstract}
%
%
\section{Introduction}
\begin{figure}[t]
    \centering
    \includegraphics[width=2.5in]{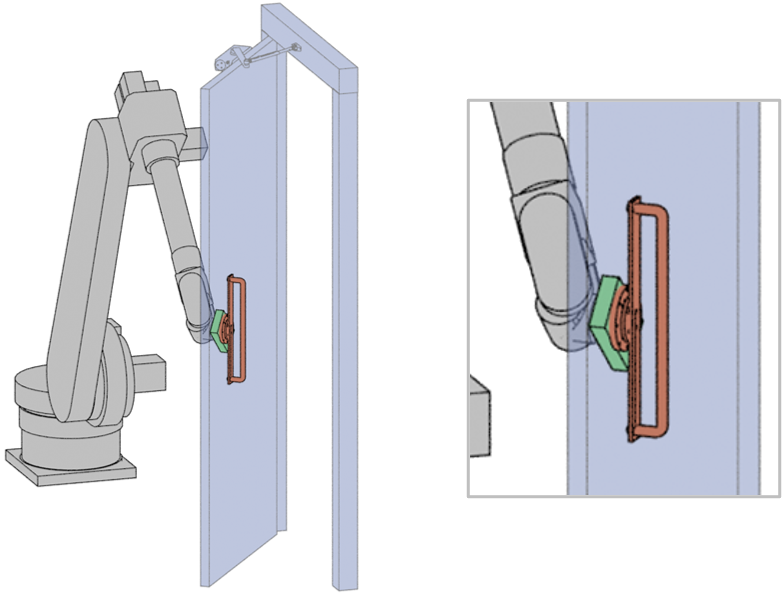}
    \caption{ An admittance-type manipulator acting as a virtual door (blue). The user applies forces through the handle end effector (red), which are measured by the sensor (green).}
    \label{fig:Fig. 1}
\end{figure}

\lettrine{H}{aptic} interfaces can reproduce the physical behavior of real-world objects, allowing users to interact with a robot as though it were the object itself. An example of this is shown in Fig.~\ref{fig:Fig. 1}. A large manipulator is equipped with a force/torque sensor (green) and a door handle end effector (red). No physical door is present. Instead, the robot reproduces the behavior of the virtual door shown in blue. The user applies forces to the handle, and the robot responds according to the model of the door's physical behavior.

While seemingly straightforward, the model required for the above example is fairly particular. First, admittance-type haptic interfaces require models of both the \textit{pre-sensor} and \textit{post-sensor} portions of a virtual object \cite{keemink2018}. In this example, the pre-sensor portion is the virtual door body, while the post-sensor portion is the door handle end effector. Second, the inertia matrix for each component must be physically consistent (symmetric and positive definite) to produce stable motion \cite{faulring2007haptic}. Accurate models of commercially available objects such as carts, doors, and appliances are rarely available and must be identified. This work presents two examples of a workflow for identifying admittance models of real-world objects with different structural constraints and levels of model complexity using the force/torque sensor onboard the haptic device, a motion-capture system, and an IMU.

This approach physically separates a given object into \textit{handle} and \textit{body} segments to identify both models separately. An adapter is made to join the handle to the force/torque sensor, which acts as an end effector. A second adapter connects the body of the object to the opposite side of the force/torque sensor. Payload estimation, a well known technique in robotics, estimates the handle inertial parameters \cite{atkeson1986}. Physically consistent inertial and friction parameters are required so the parameterizations from \cite{rucker2022} are used to guarantee this. The object is then reassembled by connecting the handle to the body with the force/torque sensor between the two. The interaction wrenches applied to the object body can now be determined with knowledge of the handle model. The body wrenches are then used with various least squares techniques to identify the body parameters.

To the best of our knowledge, this is the first complete workflow for identifying admittance models for real-world objects with various constraints and mechanisms using techniques that guarantee physical consistency of inertia and friction parameters. Models are produced for a hydraulic door closer and a wheelbarrow, representing objects of differing constraint types and model complexity. For the door, a four-bar linkage kinematic model and zone-based fluid damping model are identified. For the wheelbarrow, a full 3D model with a spherical wheel contact model and no-slip rolling is identified. This work addresses the identification problem specifically; validation in closed-loop admittance control is left to future work.

The remainder of this paper covers related work for identifying similar objects. Mathematical foundations are then applied to identify the handle models, followed by derivations and estimations of the door and wheelbarrow bodies. The results are then presented and discussed, followed by comments on future work.

\section{Related Work}

The first object is a standard interior door with an overhead hydraulic door closer. This style of door has not been the subject of system identification, but similar doors have been. The dynamics of a
refrigerator door including inertia, friction, stiffness, and magnetic
torque were modeled in \cite{shin2012}. Inertial, friction,
stiffness, thermal, and magnetic effects in both before and after gasket detachment
zones were modeled for general appliance doors such as an oven or
refrigerator in \cite{cucinotta2017}. Doors of consumer
products were reverse engineered \cite{graziosi2014} using an
optimization to determine spring stiffness and preload, door mass, door
moment of inertia, as well as torque and friction torques at various
points in the door-driving linkage. Look-up tables were used in \cite{jain2013} for door opening force vs opening angle on a freezer
door, kitchen cabinet, office cabinet, refrigerator door, and
spring-loaded door. Torque profiles were recorded for playback from car
doors and separated into active torque, such as gravity, and passive
inertial torque, like friction in \cite{ma2024}. These torque
profiles were modified and used in a haptic device in \cite{kim2025}.
In \cite{strolz2009} an extensive model of a car door is created with
look-up tables for door detent and air ventilation, gravity torque from
known geometry, as well as identified scalars for friction, rubber
packing, and end stops. This is the first identification of a complete dynamic model for this style of door and closer mechanism, guaranteeing physical consistency of inertia and friction parameters \cite{rucker2022}.

Wheelbarrow models are less studied than those of doors. Interaction forces with a wheelbarrow were recorded in \cite{luttmer2025} and a sagittal plane model of a wheelbarrow is used in \cite{vaz2021}. Single-wheel dynamics have been studied for unicycle balancing
and control \cite{han2014} \cite{cao2024}. This work is the first to identify a full 3D model of a wheelbarrow from interaction data while guaranteeing physical consistency of inertia and friction parameters \cite{rucker2022}.

\section{Methods}
The identification in this paper uses a force/torque sensor, a motion capture system, and an IMU. The sensors included a 6-DOF force/torque sensor (ATI9105-TIF-OMEGA85), a ten-camera VICON system (MXF20, MXF40 cameras) \cite{luttmer2025}, and an IMU (Parker-Lord Microstrain 3DMGX5-AR). Not every sensor was required for each identification problem. The force/torque sensor and the IMU operate at 1 kHz. Motion capture data is sampled at 100 Hz, and when used in regression, the other data is downsampled to match.

Data collection follows a two-stage procedure. For the handle, the sensor is first zeroed with no payload. The handle is then mounted and manually excited through various orientations and angular velocities. For the object body, the handle is removed, the sensor is zeroed, and the handle is reattached to the sensor. The previously identified handle model isolates the wrenches applied to the body, while the body is moved dynamically to excite the model parameters. For both handles and bodies, a training set is collected for parameter estimation, and a separate testing set of similar dynamic excitation is collected to verify model performance.

\subsection{Mathematical Foundations}
This section briefly reviews techniques described in \cite{atkeson1986} and \cite{rucker2022}, which provide a mathematical foundation for this paper. Consider a rigid body with mass $m$, center of mass vector
$\mathbf{c} \in \mathbb{R}^{3}$, and inertia matrix
$\mathbf{I} \in\mathbb{R}^{3 \times 3}$, subject to gravity $\mathbf{g} \in \mathbb{R}^{3}$. The center of
mass and inertia matrix are expressed relative to the fixed sensor
frame, which has angular velocity,
$\boldsymbol{\omega} \in \mathbb{R}^{3}$, angular acceleration,
$\dot{\boldsymbol{\omega}} \in \mathbb{R}^{3}$, and linear
acceleration, $\mathbf{a} \in \mathbb{R}^{3}$. The net wrench,
$\mathbf{w} \in \mathbb{R}^{6}$, in a
fixed sensor frame can be written linearly \cite{atkeson1986} as
\begin{equation}
    \mathbf{w} = \mathbf{A}\boldsymbol{\phi}
    \label{eq:wrench}
\end{equation}
where the inertial parameter vector,
$\boldsymbol{\phi} \in \mathbb{R}^{10}$, and body dynamics regressor,
$\mathbf{A} \in \mathbb{R}^{6 \times 10}$, are defined,
respectively, as
\begin{equation}
    \boldsymbol{\phi} = \begin{bmatrix}
        m & m\mathbf{c}^\top & I_{xx} & I_{xy} & I_{xz} & I_{yy} & I_{yz} & I_{zz}
    \end{bmatrix}^\top
    \label{eq:phi}
\end{equation}
\begin{equation}
    \mathbf{A} = \begin{bmatrix}
        \mathbf{a} - \mathbf{g} & \mathbf{S}(\dot{\boldsymbol{\omega}}) + \mathbf{S}(\boldsymbol{\omega})\mathbf{S}(\boldsymbol{\omega}) & \mathbf{0} \\
        \mathbf{0} & \mathbf{S}(\mathbf{g} - \mathbf{a}) & \mathbf{L}(\dot{\boldsymbol{\omega}}) + \mathbf{S}(\boldsymbol{\omega})\mathbf{L}(\boldsymbol{\omega})
    \end{bmatrix}
    \label{eq:A}
\end{equation}
where $\mathbf{S}\left( \mathbf{u} \right)\mathbf{v} = \mathbf{u} \times \mathbf{v}$,
and $\mathbf{L}( \cdot )$ is
\begin{equation}
    \mathbf{L}(\cdot) = \begin{bmatrix}
        \omega_x & \omega_y & \omega_z & 0         & 0         & 0         \\
        0        & \omega_x & 0        & \omega_y  & \omega_z  & 0         \\
        0        & 0        & \omega_x & 0         & \omega_y  & \omega_z
    \end{bmatrix}
    \label{eq:L}
\end{equation}
The net wrench at a given reference frame can be expressed as a summation of wrenches from various locations by transforming each wrench to the reference frame using the wrench transmission matrix,
$_{\ }^{a}\mathbf{W}_{b} \in \mathbb{R}^{6 \times 6}$ , written as
\begin{equation}
    {}^{a}\mathbf{W}_{b} = \begin{bmatrix}
        {}^{a}\mathbf{R}_{b} & \mathbf{0} \\
        \mathbf{S}({}^{a}\mathbf{r}_{a,b})\,{}^{a}\mathbf{R}_{b} & {}^{a}\mathbf{R}_{b}
    \end{bmatrix}
    \label{eq:wrench_transmission}
\end{equation}
where \(_{\ }^{a}\mathbf{R}_{b} \in \mathbb{R}^{3 \times 3}\) is the
rotation of frame \(b\) in frame \(a\), and
\(_{\ }^{a}\mathbf{r}_{a,b}\) is the vector from \(a\) to \(b\) in frame \(a\).

Solving \eqref{eq:wrench} for \(\boldsymbol{\phi}\) via least squares can yield physically inconsistent inertial
parameters. The work in \cite{rucker2022} proposes the mapping to
guarantee physically consistent results
\begin{equation}
    \boldsymbol{\phi}(\boldsymbol{\theta}) = e^{2\alpha}\begin{bmatrix}
        t_{1}^{2} + t_{2}^{2} + t_{3}^{2} + 1 \\
        t_{1}e^{d_{1}} \\
        t_{1}s_{12} + t_{2}e^{d_{2}} \\
        t_{1}s_{13} + t_{2}s_{23} + t_{3}e^{d_{3}} \\
        s_{12}^{2} + s_{13}^{2} + s_{23}^{2} + e^{2d_{2}} + e^{2d_{3}} \\
        -s_{12}e^{d_{1}} \\
        -s_{13}e^{d_{1}} \\
        s_{13}^{2} + s_{23}^{2} + e^{2d_{1}} + e^{2d_{3}} \\
        -s_{12}s_{13} - s_{23}e^{d_{2}} \\
        s_{12}^{2} + e^{2d_{1}} + e^{2d_{2}}
    \end{bmatrix}
    \label{eq:phi_theta}
\end{equation}
where the mapped inertial parameter vector,
\(\boldsymbol{\theta} \in \mathbb{R}^{10}\) is
\begin{equation}
    \boldsymbol{\theta} = \begin{bmatrix}
        \alpha & d_{1} & d_{2} & d_{3} & s_{12} & s_{23} & s_{13} & t_{1} & t_{2} & t_{3}
    \end{bmatrix}^\top
    \label{eq:theta}
\end{equation}
with dimensionless parameters for density scaling ($\alpha$), scaling ($d_1,d_2,d_3$), and shear ($s_{12},s_{23},s_{13}$). Parameters ($t_1,t_2,t_3$) are translational with length units.
For physically consistent damping and friction values, the transform
\(b_{x} = e^{\theta_{b_{x}}}\) can be used to ensure \(\hat{b}_{x}>0\) \cite{rucker2022}. 

Suppose a force/torque sensor measures
\(\mathbf{w}\) at timestep \(i\) with constant bias \(\mathbf{b} \in \mathbb{R}^{6}\) and a prior estimate
\({\widehat{\boldsymbol{\theta}}}_{\text{CAD}} \in \mathbb{R}^{10}\) exists from a source
such as CAD. The parameters \(\boldsymbol{\theta}\) and \(\mathbf{b}\) can
then be estimated using
\begin{equation}
    \min_{\boldsymbol{\theta},\mathbf{b}} \sum_{i=1}^{N} \left\| \mathbf{G}\left( \mathbf{w}^{i} - \mathbf{A}^{i}\boldsymbol{\phi}(\boldsymbol{\theta}) - \mathbf{b} \right) \right\|^{2} + \lambda^{2}\left\| \boldsymbol{\theta} - \widehat{\boldsymbol{\theta}}_{\text{CAD}} \right\|^{2}
    \label{eq:optimization}
\end{equation}
where \(\mathbf{G} \in \mathbb{R}^{6 \times 6}\) is a
diagonal residual weighting matrix used to scale residual components with different units, in this case Nm vs N.  The diagonal entries in the weighting matrix were set to \(\sigma_{\min}/\sigma_{j}\) where \(\sigma_{j}\) is the \(j\)th
component of the six calibration accuracy terms from the force/torque sensor
calibration sheet \cite{hollerbach2008}. \(\lambda\) is a damping term
that regularizes toward a known prior in the event of insufficient
data relative to a given parameter. This mirrors the payload
estimation problem in robotics \cite{atkeson1986}. Inclusion of $\mathbf{b}$ can improve results by capturing static bias such as force/torque from improper preloading of bolts during payload attachment. Using (\ref{eq:optimization})
gives \(\widehat{\boldsymbol{\theta}} \in \mathbb{R}^{10}\) which is mapped with (\ref{eq:phi_theta}) to recover
\({\widehat{\boldsymbol{\phi}}}_{wb,h}\),
\({\widehat{\boldsymbol{\phi}}}_{d,h} \in \mathbb{R}^{10}\), which are the
wheelbarrow handle and door handle inertial parameters, respectively. 

Parameters are considered identifiable if the regressor singular values $\mu_1$ and $\mu_N$ satisfy
\begin{equation}
    \kappa = \frac{\mu_{1}}{\mu_{N}} < 100
    \label{eq:kappa}
\end{equation}
where $\kappa$ is the condition number \cite{schroer1993}. Damped least squares have an effective condition number of
\begin{equation}
    \kappa_{\text{eff}} = \frac{\sqrt{\mu_{1}^2 + \lambda^2}}{\sqrt{\mu_{N}^2 + \lambda^2}}
    \label{eq:kappa_eff}
\end{equation}

This work uses motion capture markers to estimate kinematic and dynamic properties of the objects. At each
timestep the marker point cloud is aligned to a static reference
cloud using the Kabsch algorithm \cite{kabsch1976solution}, which gives the rigid transform
\(_{\ }^{W}\mathbf{T} \in \mathbb{R}^{4 \times 4}\) minimizing
root mean square deviation between the two clouds. The 
aligned world position of each marker, \({\widehat{P}}_{k} \in \mathbb{R}^{3}\), is
obtained by applying this transform to \(P_{k} \in \mathbb{R}^{3}\).
\subsection{Handles}
\label{subsec:handle}
The handles are separated from the objects and adapters to connect the handles to the force/torque sensor are designed and assembled. The inertial parameters are estimated by applying
\eqref{eq:optimization}. Angular acceleration was obtained by differentiating
the angular velocity from the IMU and using a fourth-order
Savitzky-Golay filter (101 ms window, \texttt{sgolay} in MATLAB). 

\subsection{Door}
\begin{figure}[t]
    \centering
    \includegraphics[width=2.5in]{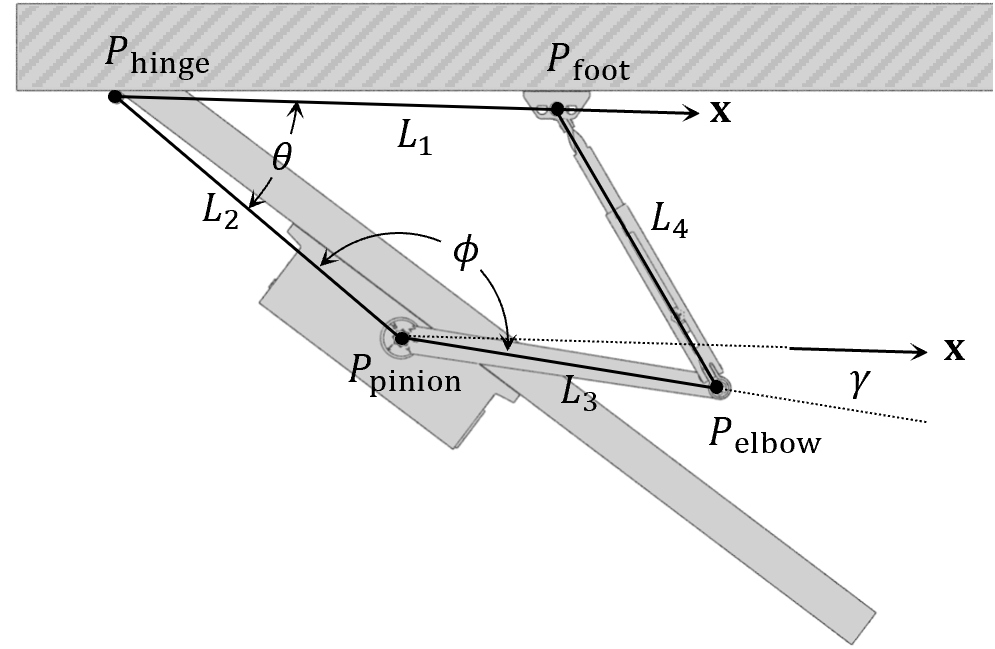}
    \caption{The kinematic model of the door-closer mechanism. The door rotates about the hinge by angle $\theta_d$, the closer pinion rotates by $\phi$ connected to a four-bar linkage of links $L_1-L_4$.}
    \label{fig:door_fbd}
\end{figure}

The door is modeled as a single rigid body rotating about a hinge,
actuated by a hydraulic door closer. Fig. ~\ref{fig:door_fbd}
shows the kinematic model where the door rotates about the hinge by
angle \(\theta_d\), while the door closer pinion rotates by \(\phi_d\).
Together, they are connected through a four-bar linkage with links
\(L_{1}\), \(L_{2}\), \(L_{3}\), and \(L_{4}\).
Freudenstein's equation \cite{norton2007}, shown below for the
four-bar linkage, can be used to solve for the absolute angle of
\(L_{3}\), \(\gamma_d\),
\begin{equation}
    K_{1}\cos\gamma_d - K_{4}\cos\theta_d + K_{5} = \cos\theta_d\cos\gamma_d + \sin\theta_d\sin\gamma_d
    \label{eq:freudenstein}
\end{equation}
which has coefficients,
\begin{gather}
    K_{1} = \frac{L_{4}}{L_{1}}, \qquad K_{4} = \frac{L_{4}}{L_{2}} \nonumber\\
    K_{5} = \frac{L_{3}^{2} - L_{4}^{2} - L_{1}^{2} - L_{2}^{2}}{2L_{1}L_{2}}
    \label{eq:K_coeffs}
\end{gather}
with quadratic coefficients
\begin{gather}
    D = \cos\theta_d - K_{1} + K_{4}\cos\theta_d + K_{5}, \qquad E = -2\sin\theta_d \nonumber\\
    F = K_{1} + (K_{4} - 1)\cos\theta_d + K_{5}
    \label{eq:DEF_coeffs}
\end{gather}
and solution
\begin{equation}
    \gamma_d = 2\arctan\left(\frac{-E \pm \sqrt{E^{2} - 4DF}}{2D}\right)
    \label{eq:gamma}
\end{equation}

\begin{figure}[t]
    \centering
    \includegraphics[width=2.5in]{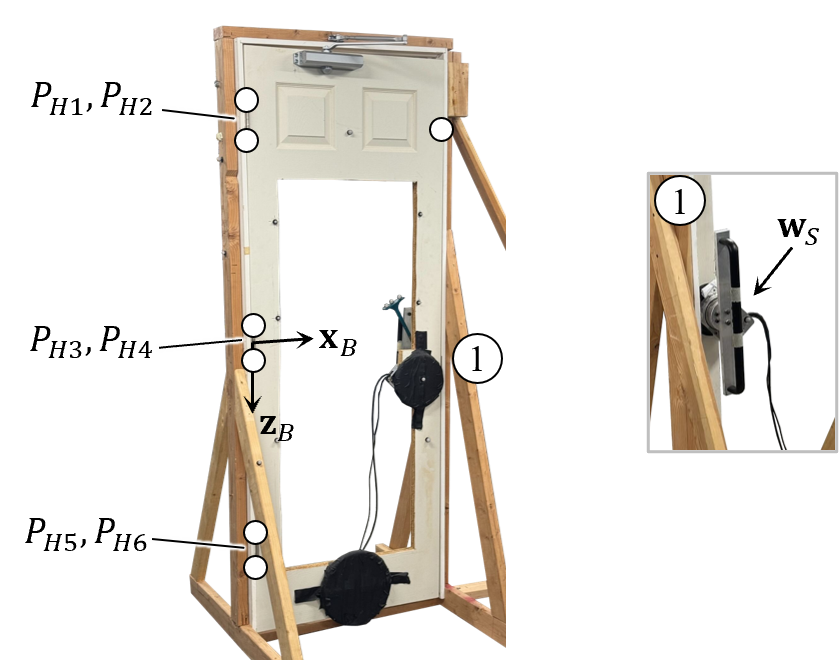}
    \caption{A 3D view of the real door showing the body coordinate frame $B$, the markers to identify the body frame, and a detailed view of the opposite side of the door where the sensor wrench, $\mathbf{w}_S$, is measured. The adapter, 1: shows a clamping assembly connecting the backside of the sensor module to the door and an aluminum bar connecting the real door handle to the sensor output flange.}
    \label{fig:door_markers}
\end{figure}

where the positive root corresponds to the physical assembly of the door
closer linkage. From the four-bar linkage geometry in
Fig. \ref{fig:door_fbd} the angle \(\phi_d\) can then be computed
as
\begin{equation}
    \phi_d = \pi + \theta_d - \gamma_d
    \label{eq:phi_fourbar}
\end{equation}

The door closer (Universal Hardware Size 1-5 Heavy Duty Commercial Door Closer, Model \#UH4031, purchased from Home Depot) is a hydraulic device containing a spring, a piston, and
valves that control the flow. The
piston is driven by the hydraulic pressure and spring forces, creating
linear rack force that acts on the output pinion, driving \(\phi_d\).
During opening (\(\dot{\phi}_d > 0\)), the fluid is free to flow
(\(\theta_d < \theta_{bc}\)) until the backcheck valve engages to prevent
overextension (\(\theta_d \geq \theta_{bc}\)). During closing
(\(\dot{\phi}_d < 0\)), the sweep valve restricts flow
(\(\theta_d \geq \theta_{l}\)) until the latch valve takes over for the
final portion of closing (\(\theta_d < \theta_{l}\)). This piecewise
behavior defines four operating zones: opening, backcheck, swing, and
latch.

\subsubsection{Physics Model}
The wrench balance at $O_B$ is
\begin{equation}
    {}^{B}\mathbf{A}_{d}\,\boldsymbol{\phi}_{d,h} = {}^{B}\mathbf{W}_{S}\left({}^{S}\mathbf{w}_{S} - {}^{S}\widehat{\mathbf{w}}_{d,h}\right) + {}^{B}\mathbf{w}_{H} + {}^{B}\mathbf{w}_{C,d} + {}^{B}\mathbf{w}_{C,s}
    \label{eq:wrench_balance_door}
\end{equation}
where \({}^{S}\mathbf{w}_{S} \in \mathbb{R}^{6}\) is the sensor frame measurement, shown in Fig. \ref{fig:door_markers}, the predicted handle wrench ${}^{S}\widehat{\mathbf{w}}_{d,h} \in \mathbb{R}^6$ is computed from the handle parameters identified in \ref{subsec:handle},
\(_{\ }^{B}\mathbf{W}_{S} \in \mathbb{R}^{6 \times 6}\) is the
transmission matrix from the sensor frame to the body frame. The hinge friction wrench ${}^{B}\mathbf{w}_{H} \in \mathbb{R}^{6}$, door closer damping wrench
${}^{B}\mathbf{w}_{C,d} \in \mathbb{R}^{6}$, and door closer spring wrench ${}^{B}\mathbf{w}_{C,s} \in \mathbb{R}^{6}$ are, respectively,
\begin{align}
    {}^{B}\mathbf{w}_{H} &= -b_{v}\dot{\theta}_d\,\mathbf{e}_6 - b_{c}\operatorname{sgn}(\dot{\theta}_d)\,\mathbf{e}_6, \label{eq:wrench_hinge}\\
    {}^{B}\mathbf{w}_{C,d} &= -\nu\tau_{b}(\dot{\phi}_d)\,\mathbf{e}_6, \label{eq:wrench_closer_damp}\\
    {}^{B}\mathbf{w}_{C,s} &= \nu\tau_{s}(\phi_d)\,\mathbf{e}_6 \label{eq:wrench_closer_spring}
\end{align}
where $\mathbf{e}_6 = [0,0,0,0,0,1]^\top$ is the sixth standard basis vector, where $b_{v}$ and $b_{c}$ are the viscous and coulomb damping coefficients at the hinge,
$\tau_{b}(\dot{\phi}_d)$ is the door closer damping torque, $\tau_{s}(\phi_d)$ is the door closer
spring torque, and $\nu = \dot{\phi}_d/\dot{\theta}_d$ is the velocity ratio \cite{norton2007}.

The door closer damping uses a laminar approximation where the torque is proportional to $\dot{\phi}_d$ \cite{pritchard2011} defined by
\begin{equation}
    \tau_{b}(\dot{\phi}_d) = \begin{cases} w_{bc}b_{bc}\dot{\phi}_d & \dot{\phi}_d > 0 \\ \left(w_{s}b_{s} + w_{l}b_{l}\right)\dot{\phi}_d & \dot{\phi}_d < 0 \end{cases}
    \label{eq:tau_b}
\end{equation}
where $b_{bc}$, $b_s$, and $b_l$ are the backcheck, sweep, and latch damping coefficients, respectively and weights of
\begin{align}
    w_{bc} &= \frac{1}{2}\left(1 + \tanh\left(\frac{\theta_d - \theta_{bc}}{\varepsilon_{bc}}\right)\right) \label{eq:w_bc}\\
    w_{l} &= \frac{1}{2}\left(1 + \tanh\left(\frac{\theta_{l} - \theta_d}{\varepsilon_{l}}\right)\right) \label{eq:w_l}\\
    w_{s} &= 1 - w_{l} \label{eq:w_s}
\end{align}
where \(\varepsilon_{l}\) and $\varepsilon_{bc}$ control the width of the smooth transition between zones using tanh. The damping and friction parameters are then grouped into $\boldsymbol{\theta}_H$ as
\begin{equation}
    \boldsymbol{\theta}_H = \begin{bmatrix} \theta_{b_{v}} & \theta_{b_{c}} & \theta_{b_{bc}} & \theta_{b_{s}} & \theta_{b_{l}} \end{bmatrix}^T \in \mathbb{R}^5
    \label{eq:theta_H}
\end{equation}
The spring torque is modeled with a polynomial as
\begin{equation}
    \tau_{\text{s}}(\phi_d) = c_{3}\phi_d^{3} + c_{2}\phi_d^{2} + c_{1}\phi_d + c_{0}
    \label{eq:tau_spring}
\end{equation}
where \(c_{0}\), \(c_{1}\), \(c_{2}\), and \(c_{3}\) are polynomial
coefficients.

Separating the identification into two steps can improve the regression behavior by creating two smaller identification problems: first, a quasi-static test identifies the spring model, $\hat{\tau}_s(\phi_d)$, and second, a dynamic test identifies the remaining inertia, damping, and friction parameters. Assuming the closer spring model has now been identified, $\hat{\mathbf{w}}_{C,s}$, the regression vectors $\mathbf{y}$ and $\mathbf{f}(\boldsymbol{\theta}, \boldsymbol{\theta}_H)$ are written as
\begin{align}
    \mathbf{y} &= {}^{B}\mathbf{W}_{S}\left({}^{S}\mathbf{w}_{S} - {}^{S}\widehat{\mathbf{w}}_{d,h}\right) + {}^{B}\hat{\mathbf{w}}_{C,s} \in \mathbb{R}^6 \label{eq:y_door_6dof}\\
    \mathbf{f}(\boldsymbol{\theta}, \boldsymbol{\theta}_H) &= {}^{B}\mathbf{A}_{d}\boldsymbol{\phi}_{d,b} - {}^{B}\mathbf{w}_{H}(\boldsymbol{\theta}_H) - {}^{B}\mathbf{w}_{C,d}(\boldsymbol{\theta}_H) \in \mathbb{R}^6 \label{eq:f_door_6dof}
\end{align}
However, the door is constrained about $\mathbf{z}_B$, so pre-multiplying $\mathbf{y}$ and $\mathbf{f}(\boldsymbol{\theta}, \boldsymbol{\theta}_H)$ by $\mathbf{e}_6$ gives scalar versions
\begin{align}
    y &= \tau_{\text{ext}} + \nu\hat{\tau}_{s}(\phi_d) \label{eq:y_door_scalar}\\
    f(\theta_I, \boldsymbol{\theta}_H) &= e^{\theta_I}\ddot{\theta}_d + e^{\theta_{b_v}}\dot{\theta}_d + e^{\theta_{b_c}}\sgn(\dot{\theta}_d) + \nu\tau_{b}(\dot{\phi}_d, \boldsymbol{\theta}_H) \label{eq:f_door_scalar}
\end{align}
where $e^{\theta_I}$ is the inertia $I$, $\dot{\theta}_d$ and $\ddot{\theta}_d$ are the hinge velocity and acceleration, and
\begin{equation}
    \tau_{\text{ext}} = \mathbf{e}_{6}^\top\,{}^{B}\mathbf{W}_{S}({}^{S}\mathbf{w}_{S} - {}^{S}\hat{\mathbf{w}}_{d,h})
    \label{eq:tau_ext_wrench}
\end{equation}
Note that $e^{\theta_I}$ is used in \eqref{eq:f_door_scalar} instead of $\boldsymbol{\phi}(\boldsymbol{\theta})$ from \eqref{eq:phi_theta} because the inertia is now only a single scalar. The final scalar minimization for the door parameters is
\begin{equation}
    \min_{\theta_I, \boldsymbol{\theta}_H} \sum_{i=1}^{N} \left(y^i - f^i(\theta_I, \boldsymbol{\theta}_H)\right)^2
    \label{eq:opt_door}
\end{equation}
which can be run using \texttt{lsqnonlin} in MATLAB. The code used is available at \cite{baum2026admittance}.

\subsubsection{Experimental Setup}
The instrumented door is shown in Fig. \ref{fig:door_fbd}. A
wooden frame was constructed to make the door free-standing and visible to the motion capture cameras. A total of 16 markers were used in the Kabsch algorithm. A large section of the door
panel was removed to improve visibility of the motion capture markers. A
weight was attached at the bottom of the door to create the feel of a
heavy door. The sensor module, which includes the IMU with force/torque sensor, is placed between the door and the door
handle.

\subsubsection{Body Kinematics}
Solving \eqref{eq:opt_door} requires $_{\ }^{B}\mathbf{R}_{S}$ and $_{\ }^{B}\mathbf{r}_{B,S}$ to find ${}^{B}\mathbf{W}_{S}$. First,
\begin{equation}
    {}^{B}\mathbf{R}_{S} = {}^{W}\mathbf{R}_{B}^\top\,{}^{W}\mathbf{R}_{S}
    \label{eq:BR_S}
\end{equation}
where \(_{\ }^{W}\mathbf{R}_{S} \in \mathbb{R}^{3 \times 3}\) is known. The vector \(_{\ }^{B}\mathbf{r}_{B,S}\) is
\begin{equation}
    {}^{B}\mathbf{r}_{B,S} = {}^{W}\mathbf{R}_{B}^\top\left({}^{W}\mathbf{r}_{W,S} - {}^{W}\mathbf{r}_{W,B}\right)
    \label{eq:r_BS}
\end{equation}
where \(_{\ }^{W}\mathbf{r}_{W,S}\) is known from the load cell markers. 

Remaining unknowns $_{\ }^{W}\mathbf{R}_{B}$ and $_{\ }^{W}\mathbf{r}_{W,B}$ can be found from motion capture and the aligned point cloud. The body frame is built using markers
\(\left\{ {\widehat{P}}_{H1},\ldots,{\widehat{P}}_{H6},{\widehat{P}}_{L} \right\}\)
as shown in Fig. \ref{fig:door_fbd}. The body frame origin,
\(_{\ }^{W}\mathbf{r}_{W,B}\), is placed at the mean of the six hinge
markers. \(\mathbf{z}_{B}\) is placed along the markers.
\(\mathbf{x}_{B}\) points from the body frame origin to
\({\widehat{P}}_{L}\) along the plane perpendicular to
\(\mathbf{z}_{B}\). \(\mathbf{y}_{B}\) is then known from the cross
product. The door angle, \(\theta_d\), is computed from the rotation of
\(\mathbf{x}_{B}\) relative to the closed position,
\(\mathbf{x}_{B,\text{closed}}\).
\subsubsection{Body Dynamics}
The door dynamics, \(\dot{\theta}_d\) and \(\ddot{\theta}_d\), are computed
from differentiating $\theta_d$ and filtering with a
fourth-order Savitzky-Golay filter (310 ms window size).

\subsubsection{Data Collection}
Quasistatic data was collected to estimate \({\widehat{\tau}}_s\)
in \eqref{eq:tau_spring}. The door was moved through the entire
opening and closing range of motion in 4 degree increments. Data outside
the threshold \(\dot{\theta}_d < 0.02\) rad/s was removed to isolate where
inertial and damping contributions are negligible. Dynamic data to estimate $(\theta_I,\boldsymbol{\theta}_H)$ in
\eqref{eq:opt_door} was collected. The door was
opened and closed many times at different speeds and through different
zones of the door's range of motion to excite the dynamics.

\subsection{Wheelbarrow}
\begin{figure}[t]
    \centering
    \includegraphics[width=\columnwidth]{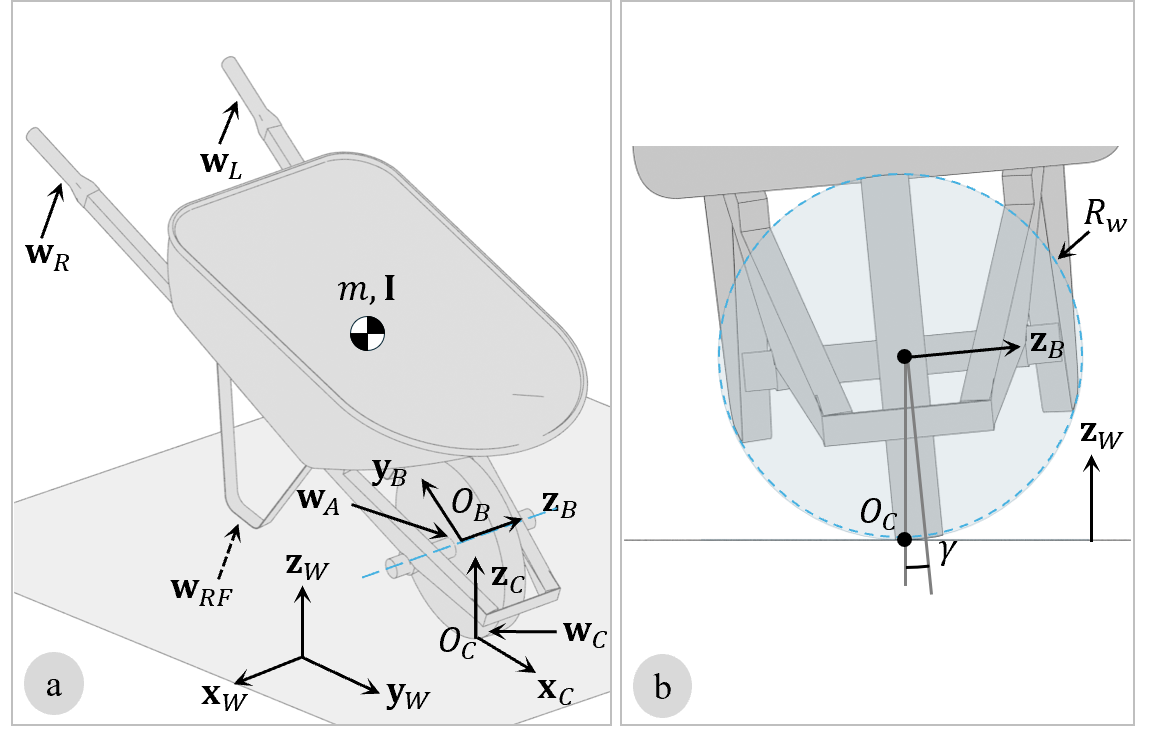}
    \caption{(a) A diagram of the wheelbarrow illustrating the body frame, $B$, the contact frame, $C$, the world frame, $W$, applied wrenches, $\mathbf{w}_R$ and $\mathbf{w}_L$, and the contact wrench, $\mathbf{w}_C$. The right foot contact wrench, $\mathbf{w}_{RF}$ is shown with a dashed line. Left foot wrench $\mathbf{w}_{LF}$ is obstructed and not labeled here. (b) The contact point model of the wheel using a sphere approximation.}
    \label{fig:wheelbarrow}
\end{figure}

A free-body diagram of the wheelbarrow (True Temper Wheelbarrow, 6 cu. ft, Model \#R6STSP14, purchased from Home Depot) is shown in
Fig.~\ref{fig:wheelbarrow} with the legs raised off the ground such that right and left foot contact wrenches, $\mathbf{w}_{RF}$ and $\mathbf{w}_{LF}$ are zero. For illustration, the visible right foot is labeled as $\mathbf{w}_{RF}$. A dashed line is used as $\mathbf{w}_{RF}$ is only present during contact which is not part of this analysis.
The body frame, with origin \(O_{B} \in \mathbb{R}^{3}\) and axes
\(\left\{ \mathbf{x}_{B},\ \ \mathbf{y}_{B},\ \ \mathbf{z}_{B} \right\} \in \mathbb{R}^{3}\),
is centered on the axle. The frame
\(\left\{ \mathbf{x}_{C},\ \ \mathbf{y}_{C},\ \ \mathbf{z}_{C} \right\} \in \mathbb{R}^{3}\)
define the contact point frame with origin \(O_{C} \in \mathbb{R}^{3}\),
where forward rolling is in the positive \(\mathbf{x}_{C}\) direction
and \(\mathbf{z}_{C}\) is normal to the ground plane. The wheel inertia is
lumped with the inertia of the bucket. Four wrenches act on the body:
the left and right handle wrenches,
\(\mathbf{w}_{L} \in \mathbb{R}^{6}\) and
\(\mathbf{w}_{R} \in \mathbb{R}^{6}\), the contact wrench
\(\mathbf{w}_{C} \in \mathbb{R}^{6}\) which is applied at the contact
point,
and the axle wrench, \(\mathbf{w}_{A} \in \mathbb{R}^{6}\), which acts
at \(O_{B}\) due to bearing friction.

The wheel is modeled in Fig.~\ref{fig:wheelbarrow}(b) with radius
\(R_{w}\) and camber angle \(\gamma_{wb}\). As the wheelbarrow leans, the
tire's real contact patch shifts laterally along the axle. To capture this shift, a spherical model is used
that projects the contact point straight down from the axle midpoint.
This approximation holds for small camber angles typical during
wheelbarrow operation (within \(\pm 7.5{^\circ}\) in this work). A rigid
disc model would not capture this lateral offset under camber.

\subsubsection{Physics Model}
The wrench balance at \(O_C\) is
\begin{multline}
    {}^{C}\mathbf{W}_{B}\,{}^{B}\mathbf{A}_{wb}\,\boldsymbol{\phi}_{wb,b} = {}^{C}\mathbf{w}_{C} + {}^{C}\mathbf{W}_{R}\left({}^{R}\mathbf{w}_{R} - {}^{R}\widehat{\mathbf{w}}_{wb,h}\right) \\
    + {}^{C}\mathbf{W}_{L}\left({}^{L}\mathbf{w}_{L} - {}^{L}\widehat{\mathbf{w}}_{wb,h}\right) + {}^{C}\mathbf{W}_{B}\,{}^{B}\mathbf{w}_{A}
    \label{eq:wrench_balance}
\end{multline}
where the predicted handle wrenches ${}^{R}\widehat{\mathbf{w}}_{wb,h}$ and ${}^{L}\widehat{\mathbf{w}}_{wb,h}$ are computed from the handle parameters identified in \ref{subsec:handle}, the axle wrench as
\begin{equation}
    {}^{B}\mathbf{w}_{A} = -b_{v}\Omega_{wb}\,\mathbf{e}_6 - b_{c}\sgn(\Omega_{wb})\,\mathbf{e}_6
    \label{eq:axle_wrench}
\end{equation}
where \(\Omega_{wb}\) is the wheel angular velocity, \(b_{v}\) is viscous
friction, and \(b_{c}\) is coulomb friction. The
contact wrench is written as
\begin{equation}
    {}^{C}\mathbf{w}_{C} = \begin{bmatrix}
        -\mu_{r}F_{N} \operatorname{sgn}\left({}^{C}\mathbf{v}_{C}^{}{}_{[1]}\right) \\
        -\mu_{r}F_{N}\operatorname{sgn}\left({}^{C}\mathbf{v}_{C}^{}{}_{[2]}\right) \\
        F_{N} \\
        0 \\
        0 \\
        -b_{\psi}\dot{\psi}_{wb}
    \end{bmatrix}
    \label{eq:contact_wrench}
\end{equation}
where \({}^{C}\mathbf{v}_{C}^{}{}_{[1]}\) and \({}^{C}\mathbf{v}_{C}^{}{}_{[2]}\) are
the \(x\) and \(y\) components of the contact point linear velocity,
\({}^{C}\mathbf{v}_{C} \in \mathbb{R}^{3}\) in the contact point
frame, \(\mu_{r}\) is the wheel rolling resistance, and \(b_{\psi}\) is
a scrubbing friction term.

The wheel velocity assuming no slip is
\begin{equation}
    \Omega_{wb} = \frac{{}^{C}\mathbf{v}_{C}{}_{[1]}}{R_{w}}
    \label{eq:omega_approx}
\end{equation}
The yaw rate, \(\dot{\psi}_{wb}\), can be determined from
\begin{equation}
    \dot{\psi}_{wb} = \left({}^{W}\mathbf{R}_{B}\,{}^{B}\boldsymbol{\omega}\right)_{[3]}
    \label{eq:psi_dot}
\end{equation}

Mass is known from a scale measurement, in which case, the mapping
\(\boldsymbol{\phi}\left( \boldsymbol{\theta} \right)\) is replaced by
\(\boldsymbol{\phi}(\boldsymbol{\theta}_{m})\) where the \(e^{2\alpha}\) term in
\eqref{eq:phi} is replaced by
\begin{equation}
    e^{2\alpha} = \frac{m_{\text{known}}}{t_{1}^{2} + t_{2}^{2} + t_{3}^{2} + 1}
    \label{eq:alpha}
\end{equation}
which reduces the unknown parameter vector to
\begin{equation}
    \boldsymbol{\theta}_{m} = \begin{bmatrix} d_{1} & d_{2} & d_{3} & s_{12} & s_{23} & s_{13} & t_{1} & t_{2} & t_{3} \end{bmatrix}^\top \in \mathbb{R}^{9}
    \label{eq:theta_m}
\end{equation}
The wheelbarrow friction parameters are grouped as
\begin{equation}
    \boldsymbol{\theta}_{f} = \begin{bmatrix} \theta_{b_{v}} & \theta_{b_{c}} & \theta_{b_{\psi}} \end{bmatrix}^\top \in \mathbb{R}^{3}
    \label{eq:theta_f}
\end{equation}
The handle wrenches transmitted to the contact frame are combined into
\(\mathbf{y} \in \mathbb{R}^{6}\) as
\begin{equation}
    \mathbf{y} = {}^{C}\mathbf{W}_{R}\left({}^{R}\mathbf{w}_{R} - {}^{R}\widehat{\mathbf{w}}_{wb,h}\right) + {}^{C}\mathbf{W}_{L}\left({}^{L}\mathbf{w}_{L} - {}^{L}\widehat{\mathbf{w}}_{wb,h}\right)
    \label{eq:y}
\end{equation}
The terms with unknown parameters in \eqref{eq:wrench_balance} are
combined into
\(\mathbf{f}\left( \boldsymbol{\theta}_{m},\boldsymbol{\theta}_{\mathbf{f}} \right) \in \mathbb{R}^{6}\)
as
\begin{multline}
    \mathbf{f}(\boldsymbol{\theta}_{m}, \boldsymbol{\theta}_{f}) = {}^{C}\mathbf{W}_{B}{}^{B}\mathbf{A}_{wb}\boldsymbol{\phi}(\boldsymbol{\theta}_{m}) \\
    - {}^{C}\mathbf{w}_{C}(\boldsymbol{\theta}_{m}, \boldsymbol{\theta}_{f}) - {}^{C}\mathbf{W}_{B}{}^{B}\mathbf{w}_{A}(\boldsymbol{\theta}_{f})
    \label{eq:f_theta}
\end{multline}
Similar to \eqref{eq:optimization} a CAD prior of
\({\widehat{\boldsymbol{\theta}}}_{m,\text{CAD}} \in \mathbb{R}^{9}\) is used. The final cost function is
\begin{equation}
    \min_{\boldsymbol{\theta}_{m}, \boldsymbol{\theta}_{f}} \sum_{i=1}^{N} \left\| \mathbf{C}\mathbf{G}\left(\mathbf{y}^{i} - \mathbf{f}^{i}(\boldsymbol{\theta}_{m}, \boldsymbol{\theta}_{f})\right) \right\|^{2} + \lambda^{2}\left\| \boldsymbol{\theta}_{m} - \widehat{\boldsymbol{\theta}}_{m,\text{CAD}} \right\|^{2}
    \label{eq:optimization_wb}
\end{equation}
where
\begin{equation}
    \mathbf{C} = \begin{bmatrix} \mathbf{0}_{3\times 3} & \mathbf{I}_{3\times 3} \end{bmatrix} \in \mathbb{R}^{3 \times 6}
    \label{eq:Pi}
\end{equation}
constrains the regression to use only the \(n_{x}\), \(n_{y}\), and \(n_{z}\)
components of the wrench balance as there is no measurement of $F_N$ from equipment such as a force plate. \eqref{eq:optimization_wb} can be solved using \texttt{lsqnonlin} in MATLAB. The code used is available at \cite{baum2026admittance}.

For completeness of the physics model, the motion capture markers also provide the vectors from the body frame
to the leg contact points \(_{\ }^{B}\mathbf{r}_{B,LF}\),
\(_{\ }^{B}\mathbf{r}_{B,RF} \in \mathbb{R}^{3}\). The left foot
penetration depth, $\delta_{LF}$, can be computed as
\begin{equation}
    \delta_{LF} = \left({}^{W}\mathbf{r}_{W,B} + {}^{W}\mathbf{R}_{B}\,{}^{B}\mathbf{r}_{B,LF}\right)_{[3]}
    \label{eq:delta_LF}
\end{equation}
which can be used in a future haptic simulation in some surface contact model $c(\delta_{LF},\dot{\delta}_{LF})$.

\subsubsection{Experimental Setup}
Fig.~\ref{fig:wb_setup} shows the experimental setup for the
wheelbarrow with the motion capture marker set with occluded markers shown in blue.
For the full wheelbarrow, sensor modules are mounted in-line with both
the left and right handles. Each sensor module carries three motion
capture markers to localize the sensor frame. Nine additional markers
were added to the wheelbarrow bucket for a more robust point cloud match. A
static 50 lb payload was secured inside the wheelbarrow bucket.

\subsubsection{Body Kinematics}
\begin{figure}[t]
    \centering
    \includegraphics[width=3in]{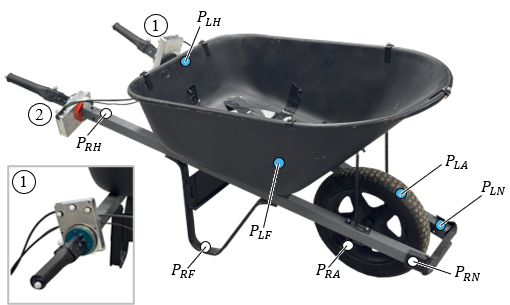}
    \caption{An image of the real wheelbarrow with markers used to identify the wheelbarrow body frame, $B$. The left (1) and right (2) sensor modules with an additional view of (1). 3D printed adapters connect the square tubing to the sensor module output flange and base.}
    \label{fig:wb_setup}
\end{figure}

Wrench transmission matrices \(_{\ }^{C}\mathbf{W}_{B}\), \(_{\ }^{C}\mathbf{W}_{L}\), and \(_{\ }^{C}\mathbf{W}_{R}\) can be computed from the markers in Fig.~\ref{fig:wb_setup}.
The aligned markers are used to compute the body frame axes in the world
frame
\begin{align}
    {}^{W}\mathbf{z}_{B} &= \frac{\widehat{P}_{LA} - \widehat{P}_{RA}}{\|\widehat{P}_{LA} - \widehat{P}_{RA}\|}, \label{eq:zB}\\
    {}^{W}\mathbf{y}_{B} &= \frac{\left(\widehat{P}_{RH} - \widehat{P}_{RN}\right) + \left(\widehat{P}_{LH} - \widehat{P}_{LN}\right)}{\left\|\left(\widehat{P}_{RH} - \widehat{P}_{RN}\right) + \left(\widehat{P}_{LH} - \widehat{P}_{LN}\right)\right\|}, \label{eq:yB}\\
    {}^{W}\mathbf{x}_{B} &= {}^{W}\mathbf{y}_{B} \times {}^{W}\mathbf{z}_{B} \label{eq:xB}
\end{align}
\begin{align}
    {}^{W}\mathbf{R}_{B} &= \begin{bmatrix} {}^{W}\mathbf{x}_{B} & {}^{W}\mathbf{y}_{B} & {}^{W}\mathbf{z}_{B} \end{bmatrix}, \label{eq:RB}\\
    {}^{W}\mathbf{r}_{W,B} &= \frac{1}{2}\left(\widehat{P}_{LA} + \widehat{P}_{RA}\right) \label{eq:pB}
\end{align}
Each sensor housing was machined with a fixed, known offset to its
markers, so the same marker set directly gives
\(_{\ }^{W}\mathbf{R}_{R}\), \(_{\ }^{W}\mathbf{R}_{L}\),
\(_{\ }^{W}\mathbf{r}_{W,R}\), and \(_{\ }^{W}\mathbf{r}_{W,L}\).

The wrench transmission matrix \(_{\ }^{C}\mathbf{W}_{R}\) requires
\(_{\ }^{C}\mathbf{R}_{R}\) and \(_{\ }^{C}\mathbf{r}_{C,R}\). First,
\begin{equation}
    {}^{C}\mathbf{R}_{R} = {}^{W}\mathbf{R}_{C}^\top\,{}^{W}\mathbf{R}_{R}
    \label{eq:CR_R}
\end{equation}
where \(_{\ }^{W}\mathbf{R}_{R} \in \mathbb{R}^{3 \times 3}\) is known
from the load cell motion capture markers. The rotation of the contact
frame, \(_{\ }^{W}\mathbf{R}_{C} \in \mathbb{R}^{3 \times 3}\) is
determined using the axle markers, which gives
\(_{\ }^{W}\mathbf{z}_{B}\)
\begin{align}
    {}^{W}\mathbf{z}_{C} &= \begin{bmatrix} 0 & 0 & 1 \end{bmatrix}^\top, \label{eq:zC}\\
    {}^{W}\mathbf{y}_{C} &= \frac{{}^{W}\mathbf{z}_{B} - \left({}^{W}\mathbf{z}_{C} \cdot {}^{W}\mathbf{z}_{B}\right){}^{W}\mathbf{z}_{C}}{\left\|{}^{W}\mathbf{z}_{B} - \left({}^{W}\mathbf{z}_{C} \cdot {}^{W}\mathbf{z}_{B}\right){}^{W}\mathbf{z}_{C}\right\|}, \label{eq:yC}\\
    {}^{W}\mathbf{x}_{C} &= {}^{W}\mathbf{y}_{C} \times {}^{W}\mathbf{z}_{C} \label{eq:xC}
\end{align}
\begin{equation}
    {}^{W}\mathbf{R}_{C} = \begin{bmatrix} {}^{W}\mathbf{x}_{C} & {}^{W}\mathbf{y}_{C} & {}^{W}\mathbf{z}_{C} \end{bmatrix}
    \label{eq:RC}
\end{equation}
the vector \(_{\ }^{C}\mathbf{r}_{C,R}\) as
\begin{equation}
    {}^{C}\mathbf{r}_{C,R} = {}^{W}\mathbf{R}_{C}^\top\left({}^{W}\mathbf{r}_{W,R} - {}^{W}\mathbf{r}_{W,C}\right)
    \label{eq:r_CR}
\end{equation}
where \(_{\ }^{W}\mathbf{r}_{W,R}\) is known from the load cell markers
and
\begin{equation}
    {}^{W}\mathbf{r}_{W,C} = \big[\,{}^{W}\mathbf{r}_{W,B}{}_{[1]},\ {}^{W}\mathbf{r}_{W,B}{}_{[2]},\ 0\,\big]^\top
    \label{eq:pC}
\end{equation}
From symmetry, the derivation for \(_{\ }^{C}\mathbf{R}_{L}\) and
\(_{\ }^{C}\mathbf{r}_{C,L}\) is identical only with \(L\) subscripts.
Next, computing matrix \(_{\ }^{C}\mathbf{W}_{B}\) requires
\(_{\ }^{C}\mathbf{R}_{B}\) and \(_{\ }^{C}\mathbf{r}_{C,B}\). First,
\begin{equation}
    {}^{C}\mathbf{R}_{B} = {}^{W}\mathbf{R}_{C}^\top\,{}^{W}\mathbf{R}_{B}
    \label{eq:CR_B}
\end{equation}
where
\begin{equation}
    {}^{W}\mathbf{R}_{B} = \begin{bmatrix} {}^{W}\mathbf{x}_{B} & {}^{W}\mathbf{y}_{B} & {}^{W}\mathbf{z}_{B} \end{bmatrix}
    \label{eq:RB_wb}
\end{equation}
where \(_{\ }^{W}\mathbf{y}_{B}\) and \(_{\ }^{W}\mathbf{z}_{B}\) are
known directly from motion capture.

\subsubsection{Body Dynamics}
Frame \(B\) is now fully known from motion capture. The frame is differentiated and a fourth order Savitzky-Golay filter with a 110
ms window was used for accelerations and velocities. 

\subsubsection{Data Collection}
Body estimation data was collected by first removing the handles and
zeroing out the sensors. Data was recorded while a human interacted with the wheelbarrow with the intention of sufficiently exciting the dynamics by: pushing forward/backwards, rolling side to side,
yawing to scrub the tire on the ground, and lifting the bucket up/down
to simulate dumping. 
%

\section{Results and Discussion}

All estimated parameters are reported in Table~\ref{tab:all_params}. The inertia and all damping parameters are positive and physically consistent. All effective condition numbers, $\kappa_{\text{eff}}$, were less than 100 (Table \ref{tab:all_params}), with $\lambda$ increased for each regression until this threshold was met. The values of $\lambda$ are reasonable as the magnitudes of the singular values happen to be large.
\subsubsection{Handles}
The handle estimation results are reported in Table~\ref{tab:all_params}. The maximum RMS residuals of 0.64 N and 0.042 Nm are sufficiently low. For context, the RMS error relative to the signal range is reported as a percentage. All values are $<3\%$ of the signal range except the $n_z$ component for the wheelbarrow handle at 13.7$\%$. This direction is along the length of the cylindrical handle, which is difficult to excite manually. This workflow relies on manual excitation of the object dynamics, so parameters that are difficult to excite by hand will benefit from a physically consistent CAD prior, such as the solid cylinder approximation used for the handles.

\begin{figure}[t]
    \centering
    \includegraphics[width=\columnwidth]{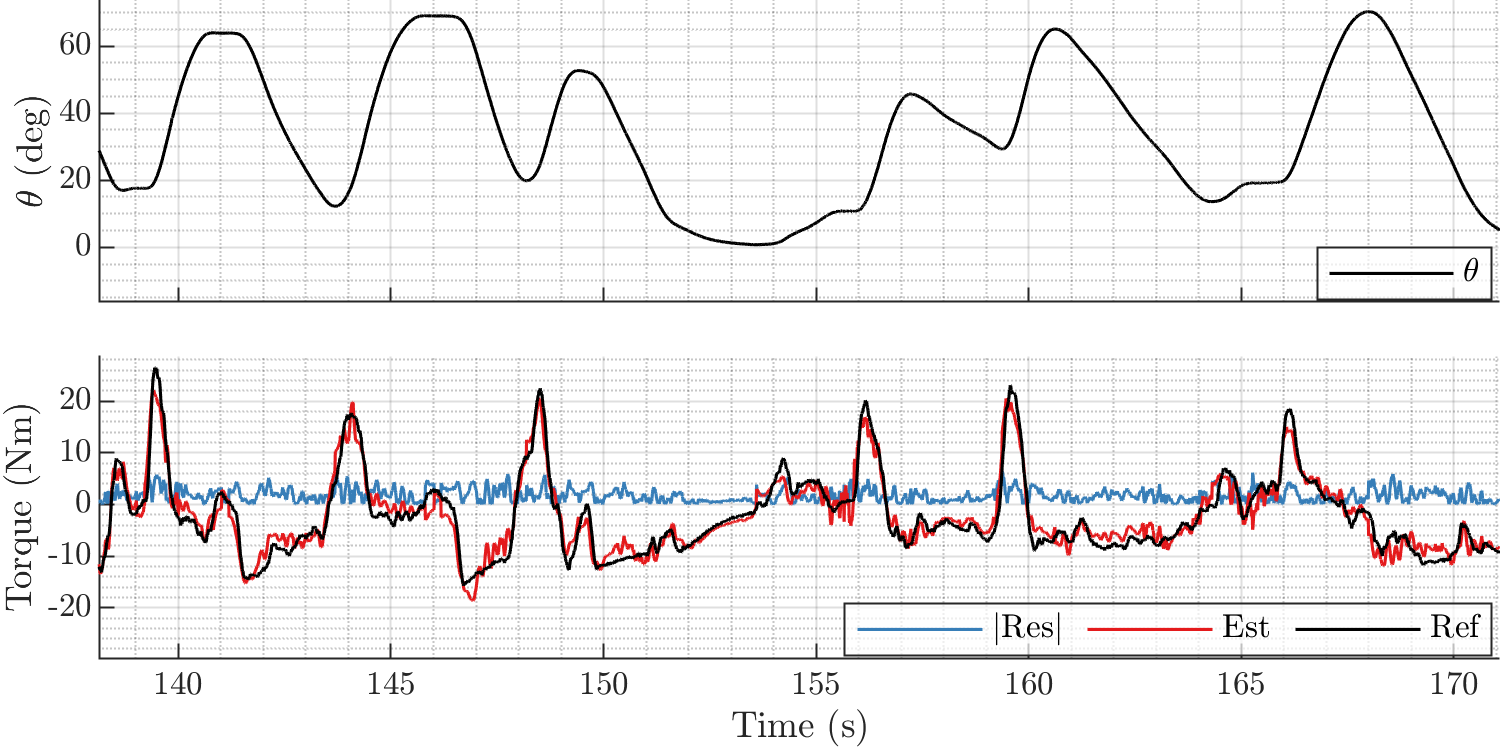}
    \caption{Door test set results. (a) Door angle $\theta_d$. (b) Reference door hinge torque, $y$ \eqref{eq:y_door_scalar} vs estimated door hinge torque, $f(\hat{\theta}_I,\hat{\boldsymbol{\theta}}_H)$ \eqref{eq:f_door_scalar}.}
    \label{fig:door_results}
\end{figure}
\begin{figure}[t]
    \centering
    \includegraphics[width=\columnwidth]{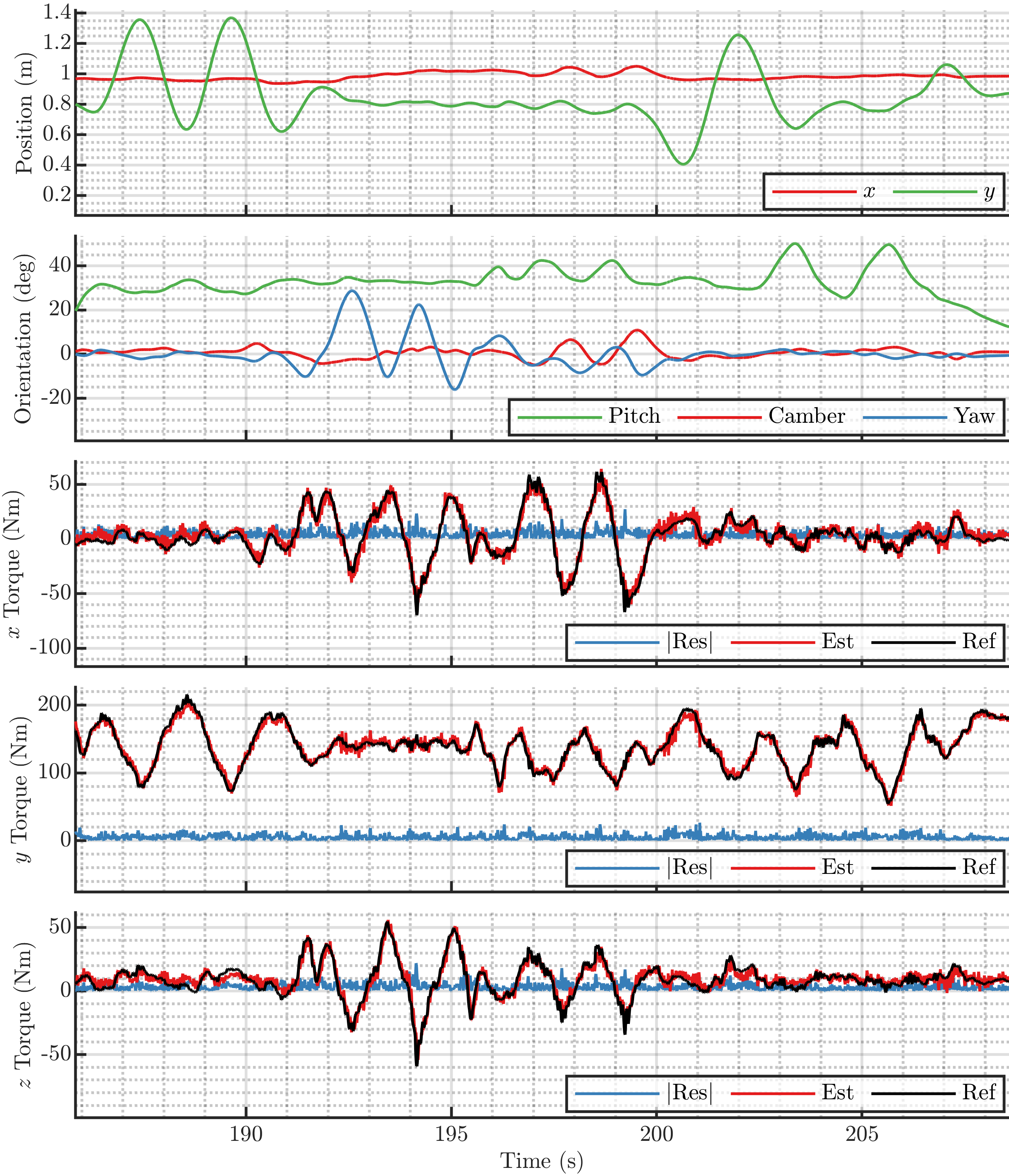}
    \caption{Wheelbarrow test set results. (a) Contact point position ${}^W\mathbf{r}_{W,C[1]}$ and ${}^W\mathbf{r}_{W,C[2]}$. (b) Pitch $\theta_{wb}$, camber $\gamma_{wb}$, and yaw $\psi_{wb}$. Residual components (c) $\tau_x$, (d) $\tau_y$, and (e) $\tau_z$ between $\mathbf{y}$ \eqref{eq:y} and $\mathbf{f}(\hat{\boldsymbol{\theta}}_m, \hat{\boldsymbol{\theta}}_f)$ \eqref{eq:f_theta}.}
    \label{fig:wb_results}
\end{figure}

\begin{table}[t]
\caption{Estimated parameters for all models.}
\label{tab:all_params}
\centering
\footnotesize
\begin{tabular}{l l cccc}
\hline
 & & \multicolumn{1}{c}{WB} & \multicolumn{1}{c}{WB} & \multicolumn{1}{c}{Door} & \multicolumn{1}{c}{Door} \\
 & & \multicolumn{1}{c}{Handle} & \multicolumn{1}{c}{Body} & \multicolumn{1}{c}{Handle} & \multicolumn{1}{c}{Body} \\
\hline
\multicolumn{6}{l}{\textit{Inertial - CAD prior in italics}} \\
$m$        & [kg]        & 0.856    & 45.0      & 1.571      & --        \\
           &             & \textit{0.900}   & \textit{45.0}    & \textit{1.500}   &           \\
$c_x$      & [m]         & 0.0001    & $-$0.115  & -0.001      & --        \\
           &             & \textit{0}       & \textit{$-$0.112} & \textit{0}      &           \\
$c_y$      & [m]         & 0.0001    & 0.445     & $-$0.001   & --        \\
           &             & \textit{0}       & \textit{0.450}   & \textit{0}       &           \\
$c_z$      & [m]         & 0.077    & 0.012     & 0.025      & --        \\
           &             & \textit{0.075}   & \textit{0.0004}  & \textit{0.02}       &           \\
$I_{xx}$   & [kg\,m$^2$] & 0.0126  & 13.56     & 0.0077    & --        \\
           &             & \textit{0.0068} & \textit{12.771}  & \textit{0.0064} &           \\
$I_{yy}$   & [kg\,m$^2$] & 0.0124  & 2.80      & 0.0157    & --        \\
           &             & \textit{0.0068} & \textit{1.658}   & \textit{0.0065} &           \\
$I_{zz}$   & [kg\,m$^2$] & 0.0004  & 12.55     & 0.0180    & 6.53      \\
           &             & \textit{0.0002} & \textit{12.805}  & \textit{0.0114} &           \\
$I_{xy}$   & [kg\,m$^2$] & 0.0001  & 2.63      & $-$0.0078 & --        \\
           &             & \textit{0}       & \textit{2.324}   & \textit{0.0055} &           \\
$I_{xz}$   & [kg\,m$^2$] & 0.0002  & 0.28      & $-$0.0002 & --        \\
           &             & \textit{0}       & \textit{0.007}   & \textit{0}       &           \\
$I_{yz}$   & [kg\,m$^2$] & 0.0003  & 0.61      & 0.0006    & --        \\
           &             & \textit{0}       & \textit{$-$0.007} & \textit{0}      &           \\
\hline
\multicolumn{6}{l}{\textit{Damping}} \\
$b_{bc}$   & [Nm\,s/rad] &          &           &            & 7.52     \\
$b_s$      & [Nm\,s/rad] &          &           &            & 4.47      \\
$b_l$      & [Nm\,s/rad] &          &           &            & 2.57      \\
\hline
\multicolumn{6}{l}{\textit{Friction}} \\
$b_v$      & [Nm\,s/rad] &          & 0.388     &            & 4.75      \\
$b_c$      & [Nm]        &          & 1.172     &            & 1.79      \\
$b_\psi$   & [Nm\,s/rad] &          & 2.943     &            &           \\
\hline
\multicolumn{6}{l}{\textit{Stiffness}} \\
$c_0$      & [Nm]         &         &           &            & 18.01      \\
$c_1$      & [Nm/rad]     &         &           &            & $-$17.10  \\
$c_2$      & [Nm/rad$^2$] &         &           &            & 5.34    \\
$c_3$      & [Nm/rad$^3$] &         &           &            & $-$0.70 \\
\hline
\multicolumn{6}{l}{\textit{Regression - $^*$effective $\kappa$}} \\
Train  & [sec]        & 27       & 30       & 45         & 57          \\
Test  & [sec]        & 16       & 23     & 19         & 33          \\
$\lambda$  & [--]        & 10       & 100       & 20         &           \\
$\kappa$   & [--]        & 23.9$^*$     & 97.5$^*$ & 27.5$^*$       & 15.0  \\
\hline
\multicolumn{6}{l}{\textit{RMS residuals - \% is percent of signal range}} \\
$f_x$      & [N]   & 0.35  &   & 0.46  &      \\
           & [\%]  & \textit{1.6}   &  & \textit{1.4}   &      \\
$f_y$      & [N]   & 0.46  &       & 0.64  &      \\
           & [\%]  & \textit{1.6}   &       & \textit{1.5}   &      \\
$f_z$      & [N]   & 0.38  &       & 0.40  &      \\
           & [\%]  & \textit{3.0}   &       & \textit{1.7}   &      \\
$n_x$      & [Nm]  & 0.023 &  5.31 & 0.022 &      \\
           & [\%]  & \textit{0.8}   & \textit{4.0} & \textit{1.3}   &      \\
$n_y$      & [Nm]  & 0.020 &  6.77 & 0.031 &      \\
           & [\%]  & \textit{0.8}   & \textit{4.2} & \textit{1.5}   &      \\
$n_z$      & [Nm]  & 0.014 &  4.42 & 0.042 & 2.19 \\
           & [\%]  & \textit{13.7}  & \textit{3.9} & \textit{1.0}   & \textit{5.2} \\
\end{tabular}
\end{table}

\subsubsection{Door}
The door model and trajectory is shown in Fig. \ref{fig:door_results}. The RMS residual is 2.19 Nm, (5.2$\%$ of the torque reference range) and the average marker error was 1.54 mm.  Parameters of $\theta_{l}=25^\circ$, $\theta_{bc}=69^\circ$, $\varepsilon_{bc}=\varepsilon_{l}=2^\circ$ were tuned empirically. 
Lumping these parameters into the identification resulted in poor conditioning or complete loss of identifiability due to linear combinations of parameters. Interestingly, $\theta_{l}$ values closer to $25^\circ$, rather than the expected $10^\circ$, gave better results which points towards a more advanced damping model could be beneficial in the future. Additionally, a turbulent version of the closer damping model $\tau_b({\phi_d})$ \eqref{eq:tau_b} was also tested, replacing $\dot{\phi}_d$ with $\dot{\phi}_d^{2}$, which is analogous to turbulent flow \cite{pritchard2011}, but resulted in higher RMS residuals.

During the quasi-static estimation of the closer spring model, $\tau_s(\phi_d)$, some hysteresis was observable in the force vs. displacement plot. The polynomial model uses a single set of coefficients fit to data from both opening and closing trajectories. A next step would be to fit independent coefficients for opening and closing.

\subsubsection{Wheelbarrow}

The wheelbarrow test results is shown in Fig. \ref{fig:wb_results} with all RMS residuals below $4.2\%$ of the signal range and an average marker error of 2.1 mm. The $x$ and $y$ components of the contact point in the world frame, ${}^W \mathbf{r}_{W,C}$ are shown as well as wheelbarrow pitch $\theta_{wb}$ (incline of ${}^{W}\mathbf{y}_{B}$), camber $\gamma_{wb}$ (lean of ${}^{W}\mathbf{z}_{B}$), and yaw $\psi_{wb}$ (heading of ${}^{W}\mathbf{x}_{C}$). The reference is not a direct measurement of the contact wrench, but a projection of the handle interaction wrenches to the contact point using the estimated kinematics. The spherical wheel approximation appears valid according to the model results. However, the field of tire dynamics is well studied and the model could be improved at the cost of measuring tire slip, which was not possible with the instrumentation used here.

The wheelbarrow solution uses the torques at the contact point to eliminate the need for a normal force measurement. This effectively lumps $\mu_r$ into the axle coulomb damping, $b_c$. The estimated value for $b_c$ was 1.17 Nm. Using the geometry of the 0.197 m radius of the wheel and the 45 kg wheelbarrow mass, 1.17 Nm would be equivalent to a $\mu_r$ of 0.013 assuming all friction was due to rolling. The true $\mu_r$ is likely between 0 and 0.013, left to be chosen in that range when simulating the admittance model.

\subsubsection{Dynamics and Object Compliance}
Acceleration and velocity data used in these identification problems can come from the IMU, differentiation of the motion capture data, or a mixture of both. The handle estimations performed well using the IMU measurements directly. The larger bodies both showed better results using kinematic quantities derived from motion capture, differentiated and filtered with a Savitzky-Golay filter. The lower performance of the IMU-based approach may have resulted from compounded errors in estimating the sensor frame.

This workflow requires adapters to mount the sensor module between the handle and the body, creating a lower bound on the applicable object mass. The instrumentation and adapters (1--2 kg in this work) could noticeably affect interaction with lighter objects such as a cane. The door and wheelbarrow have sufficient mass to make this effect negligible. The adapter assembly can also introduce compliance into the system. The handle, sensor module, and body are assumed to form a rigid chain so any residual compliance manifests as additional error.

\subsubsection{Future Work}

Future work includes implementing the identified models in the control loop of a real admittance-type robot and evaluating their performance against real objects in a subject study. Adapter compliance could also be addressed by either modeling it or improving the rigidity of the adapter assembly in future designs. Future work could address remaining modeling challenges such as hysteresis and complex fluid effects in the door closer model. 

This workflow is general and could be applied to other constrained objects such as appliances, vehicle doors, or industrial equipment, provided the object dynamics can be sufficiently excited manually and a rigid sensor mounting is feasible.

\section{Conclusion}
The proposed methodology demonstrates that physically consistent admittance models can be identified for large, constrained real-world objects without joint-level instrumentation or detailed CAD models, using only a force/torque sensor, motion capture, and an IMU. Models were successfully produced for the wheelbarrow and door, with all inertial and friction estimates guaranteed physically consistent.

\addtolength{\textheight}{-12cm}   





\bibliographystyle{IEEEtran}
\bibliography{refs}

@article{atkeson1986,
  author  = {Atkeson, Christopher G. and An, Chae H. and Hollerbach, John M.},
  title   = {Estimation of Inertial Parameters of Manipulator Loads and Links},
  journal = {The International Journal of Robotics Research},
  volume  = {5},
  number  = {3},
  year    = {1986}
}

@article{cao2024,
  author  = {Cao, Xincheng and Bui, Dang Cong and Tak\'{a}cs, D\'{e}nes and Orosz, G\'{a}bor},
  title   = {Autonomous Unicycle: Modeling, Dynamics, and Control},
  journal = {Multibody System Dynamics},
  volume  = {61},
  number  = {1},
  year    = {2024}
}

@article{faulring2007haptic,
  title={Haptic display of constrained dynamic systems via admittance displays},
  author={Faulring, Eric L and Lynch, Kevin M and Colgate, J Edward and Peshkin, Michael A},
  journal={IEEE Transactions on Robotics},
  volume={23},
  number={1},
  pages={101--111},
  year={2007},
  publisher={IEEE}
}

@incollection{hollerbach2008,
  author    = {Hollerbach, John and Khalil, Wisama and Gautier, Maxime},
  title     = {Model Identification},
  booktitle = {Springer Handbook of Robotics},
  editor    = {Siciliano, Bruno and Khatib, Oussama},
  chapter   = {14},
  pages     = {321},
  publisher = {Springer},
  year      = {2008}
}

@article{kabsch1976solution,
  title={A solution for the best rotation to relate two sets of vectors},
  author={Kabsch, Wolfgang},
  journal={Foundations of Crystallography},
  volume={32},
  number={5},
  pages={922--923},
  year={1976},
  publisher={International Union of Crystallography}
}

@incollection{schroer1993,
  author    = {Schr\"{o}er, Klaus},
  title     = {Theory of Kinematic Modelling and Numerical Procedures for Robot Calibration},
  booktitle = {Robot Calibration},
  editor    = {Bernhardt, R. and Albright, S. L.},
  publisher = {Chapman and Hall},
  address   = {London},
  year      = {1993},
  pages     = {157--196}
}

@article{cucinotta2017,
  author  = {Cucinotta, Filippo and Sfravara, Felice},
  title   = {An Appliance Door Virtual Modeling For User Experience Design},
  journal = {International Journal of Applied Engineering Research},
  volume  = {12},
  number  = {14},
  year    = {2017}
}

@article{graziosi2014,
  author  = {Graziosi, Serena and Ferrise, Francesco and Furtado, Guilherme Phillips and Bordegoni, Monica},
  title   = {Reverse Engineering of Interactive Mechanical Interfaces for Product Experience Design},
  journal = {Virtual and Physical Prototyping},
  volume  = {9},
  number  = {2},
  year    = {2014}
}

@article{han2014,
  author  = {Han, Seong I. and Lee, Jang M.},
  title   = {Balancing and Velocity Control of a Unicycle Robot Based on the Dynamic Model},
  journal = {IEEE Transactions on Industrial Electronics},
  volume  = {62},
  number  = {1},
  year    = {2014}
}

@article{jain2013,
  author  = {Jain, Advait and Kemp, Charles C.},
  title   = {Improving Robot Manipulation with Data-Driven Object-Centric Models of Everyday Forces},
  journal = {Autonomous Robots},
  volume  = {35},
  number  = {2},
  year    = {2013}
}

@article{keemink2018,
  author  = {Keemink, Arvid Q. L. and Van der Kooij, Herman and Stienen, Arno H. A.},
  title   = {Admittance Control for Physical Human--Robot Interaction},
  journal = {The International Journal of Robotics Research},
  volume  = {37},
  number  = {11},
  year    = {2018}
}

@inproceedings{kim2025,
  author    = {Kim, Ji-Sung and Ma, Jihyeong and Kyung, Ki-Uk},
  title     = {Hybrid Haptic Device for Car Door Interactions: User Perception of Torque Profiles},
  booktitle = {2025 IEEE World Haptics Conference (WHC)},
  year      = {2025}
}

@misc{baum2026admittance,
  author       = {Baum, Nathan I.},
  title        = {System Identification of Admittance Models for Large Real-World Objects},
  howpublished = {GitHub repository},
  year         = {2026},
  note         = {Available: \url{https://github.com/nate-baum/large-objects-sys-id}. Accessed: Aug. 30, 2026}
}

@article{luttmer2025,
  author  = {Luttmer, Nathaniel G. and Baum, Nathan I. and Flores-Gonzalez, Josue and Hollerbach, John M. and Minor, Mark A.},
  title   = {The Utah Manipulation and Locomotion of Large Objects ({MeLLO}) Data Library},
  journal = {Bioengineering},
  volume  = {12},
  number  = {3},
  year    = {2025}
}

@article{ma2024,
  author  = {Ma, Jihyeong and Kim, Ji-Sung and Kyung, Ki-Uk},
  title   = {A Hybrid Haptic Device for Virtual Car Door Interactions: Design and Implementation},
  journal = {IEEE Robotics and Automation Letters},
  volume  = {9},
  number  = {10},
  year    = {2024}
}

@book{norton2007,
  author    = {Norton, Robert L. and Han, Jianyou},
  title     = {Design of Machinery},
  edition   = {4},
  publisher = {McGraw-Hill},
  year      = {2007}
}

@book{pritchard2011,
  author    = {Pritchard, Philip J.},
  title     = {Introduction to Fluid Mechanics},
  edition   = {8},
  publisher = {Wiley},
  year      = {2011}
}

@article{rucker2022,
  author  = {Rucker, Caleb and Wensing, Patrick M.},
  title   = {Smooth Parameterization of Rigid-Body Inertia},
  journal = {IEEE Robotics and Automation Letters},
  volume  = {7},
  number  = {2},
  year    = {2022}
}

@inproceedings{shin2012,
  author    = {Shin, Sunghwan and Lee, In and Lee, Hojin and others},
  title     = {Haptic Simulation of Refrigerator Door},
  booktitle = {2012 IEEE Haptics Symposium (HAPTICS)},
  year      = {2012}
}

@inproceedings{strolz2009,
  author    = {Strolz, Michael and Ehinger, Claudia and Buss, Martin},
  title     = {Design of a Device for the High-Fidelity Haptic Rendering of Rotatory Car Doors},
  booktitle = {2009 2nd Conference on Human System Interactions},
  year      = {2009}
}

@inproceedings{vaz2021,
  author    = {Vaz, Jean Chagas and Torres-Reyes, Norberto and Oh, Paul Y.},
  title     = {Humanoid Interaction with Material-Moving Carts and Wheelbarrows},
  booktitle = {2021 IEEE International Symposium on Safety, Security, and Rescue Robotics (SSRR)},
  year      = {2021}
}

\end{document}